\documentclass[letterpaper, 10pt, conference]{ieeeconf}

\IEEEoverridecommandlockouts                              % This command is only needed if 
\usepackage{graphicx}
\usepackage{url}
\usepackage[export]{adjustbox} % for valign
\usepackage{xcolor}
\usepackage[ruled,vlined,norelsize,linesnumbered]{algorithm2e}
\usepackage{cite} % For ranges of refs
\usepackage{amsmath}
\usepackage{amssymb}
\let\labelindent\relax
\usepackage{enumitem}
\usepackage{tabu}

\makeatletter
\newcommand{\removelatexerror}{\let\@latex@error\@gobble}
\makeatother

\SetCommentSty{mycommfont}

\newcommand{\puckVariant}[0]{ \ensuremath{\text{PUCK\_MASK}} }
\newcommand{\alignVariant}[0]{ \ensuremath{\text{ALIGN\_MASK}} }

\newcommand{\alignVariantSTOP}[0]{ \ensuremath{\text{ALIGN\_MASK}}}

\newcommand{\LLL}[0]{ \ensuremath{L} }
\newcommand{\CCC}[0]{ \ensuremath{C} }
\newcommand{\RRR}[0]{ \ensuremath{R} }
\newcommand{\leftPuck}[0]{ \ensuremath{\text{puck}_\text{left}} }
\newcommand{\leftRobot}[0]{ \ensuremath{\text{robot}_\text{left}} }
\newcommand{\rightRobot}[0]{ \ensuremath{\text{robot}_\text{right}} }

\newcommand{\omegaMax}[0]{ \ensuremath{\omega_{max}} }
\newcommand{\vMax}[0]{ \ensuremath{v_{max}} }

\newcommand{\etal}[0]{\textit{et\,al. }}

\title{\LARGE \bf
A Swarm of Simple Robots Constructing Planar Shapes
}

\author{Andrew Vardy$^{1}$% <-this % stops a space
and Dalia S. Ibrahim$^{2}$% <-this % stops a space
\thanks{*Supported by Natural Sciences and Engineering Research (NSERC)}% <-this % stops a space
\thanks{$^{1}$Andrew Vardy is with the Departments of Computer Science and Electrical \& Computer Engineering,
        Memorial University of Newfoundland, St.\ John's, Canada 
        {\tt\small av@mun.ca}}%
\thanks{$^{2}$Dalia S. Ibrahim is with the Departments of Computer Science,
        Memorial University of Newfoundland, St.\ John's, Canada 
        {\tt\small dsibrahim@mun.ca}}%
}

\begin{document}

\maketitle
\thispagestyle{empty}
\pagestyle{empty}

%%%%%%%%%%%%%%%%%%%%%%%%%%%%%%%%%%%%%%%%%%%%%%%%%%%%%%%%%%%%%%%%%%%%%%%%%%%%%%%%
\begin{abstract}
    We present a new version of our previously proposed algorithm enabling a swarm of robots to construct a desired shape from objects in the plane.  We also describe a hardware realization for this system which makes use of simple and readily sourced components.  We refer to the task as planar construction which is the gathering of ambient objects into some desired shape.  As an example application, a swarm of robots could use this algorithm to not only gather waste material into a pile, but shape that pile into a line for easy collection.  The shape is specified by an image known as the scalar field.  The scalar field serves an analogous role to the template pheromones that guide the construction of complex natural structures such as termite mounds. In addition to describing the algorithm and hardware platform, we develop some performance insights using a custom simulation environment and present experimental results on physical robots.
\end{abstract}

%%%%%%%%%%%%%%%%%%%%%%%%%%%%%%%%%%%%%%%%%%%%%%%%%%%%%%%%%%%%%%%%%%%%%%%%%%%%%%%%
\section{INTRODUCTION}

A swarm of simple robots able to manipulate objects in their environment, configuring them into a desired shape has potential applications in cleaning, waste collection, recycling and in construction.  So far, collective robot construction has most often considered the use of specially designed materials, paired with custom robot hardware \cite{petersen2019review}.  But we believe there is great potential in using readily-sourced components and available robot platforms to build swarms that can manipulate objects in their environment.  Our approach requires no exotic hardware and incurs very little computational cost.  To specify the shape to be constructed we take inspiration from the social insects which make use of pheromones to specify the shape of their nest structure, making use of self-organizing processes during its formation \cite{camazine2003self}.  Ladley and Bullock proposed a model whereby the stationary queen termite exudes a \emph{template} pheromone causing workers to deposit material at a characteristic distance from the queen, eventually leading to the constructing of a ``royal chamber'' \cite{ladley05logistic}.  We are interested in the intersection of the capabilities of simple readily-sourced robots and the possibilities inherent in a signal playing the role of template pheromone to construct particular shapes from objects in the environment.

We also draw inspiration from the study of object clustering, pioneered as a concept by Deneubourg \etal \cite{deneubourg90sorting} and demonstrated in real-world robots by a number of researchers (see \cite{vardy14cache} for a review).  We are particularly inspired by a paper from Gauci \etal which demonstrated an extremely simple object clustering algorithm discovered via evolutionary search \cite{gauci14clustering}.  This algorithm drives each robot toward the periphery of the objects within the environment.  The robots then begin to encircle the objects while bumping against them, thus shifting them inwards incrementally.  We previously combined this approach with the template pheromone concept and proposed an algorithm called \emph{orbital construction} (OC) \cite{vardy2018orbital}.  %In the original OC algorithm
In this paper we propose a new version of this algorithm which is more flexible
and accounts for the presence of other robots which may block movement.  We also
present a hardware platform that will allow us to demonstrate the algorithm's performance.  

The OC algorithm, like Gauci \etal's, is purely reactive and is therefore easy
to describe and implement.  It also involves very few parameters which need to
be tuned.  We previously proposed a somewhat more complex algorithm for planar
construction which relies upon a set of distinct landmarks to specify the shape
of the desired structure \cite{vardy2019landmark}.  In addition, we have
investigated the use of reinforcement learning for the planar construction
problem
\cite{strickland19rl}.  Using the scalar field as part of the state space proved
to be highly successful and allowed for very efficient learning of an effective
policy.  We have also studied the allocation of different roles to different
agents and found that a local communication strategy outperformed global
communication \cite{ibrahim19adaptive}.

%The approach taken in this paper combines the template pheromone with the attractive properties of Gauci \etal's algorithm.  The template will be provided to a set of robots operating on a work surface, by painting or projecting a scalar field directly to this surface.  Using simple photosensitive sensors, the robots can determine both the value of the scalar field and its gradient.  Our hope is to use the arrays of infrared emitter-detector pairs commonly used for line following on various low-cost robots.  Similar to Gauci \etal our approach also requires a camera to detect the presence of other robots, the walls of the environment, and objects (pucks).  The goal is to move a set of randomly positioned pucks to form an enclosure with a desired shape.  We demonstrate construction of the following exemplary shapes: circular, oblong, and cross-shaped enclosures.  Since our algorithm guides robots in a clockwise orbit around the growing structure we refer to it as \textbf{orbital construction}.

% ANY GOOD?
%There are several ways of characterizing and defining the field of swarm robotics \cite{brambilla13review}.  One One the two most germaine  As an approach to the design of useful multi-robot systems, swarm robotics can be characterized by the strictly local interactions between the robots and between the robots and their environment.  

% A BETTER START?
The planar construction problem can be considered a sub-area of collective robotic construction, which was recently reviewed in \cite{petersen2019review}. Planar construction is the formation of a desired two-dimensional structure from ambient objects in the environment.  This problem entails a combination of the discovery and transport of objects, as well as manipulating these objects into a desired shape.  As such, the problem is related to work on foraging \cite{ostergaard01bucket,shell06foraging,lein08adaptive}, clustering objects of a single type \cite{deneubourg90sorting,beckers94stigmergy,maris96heap,martinoli99probabilistic,kazadi02convergence,vardy14cache}, sorting objects of different types  
\cite{melhuish98collective,melhuish01patch,wang03sorting,verret04sorting,
melhuish06clustering} and the construction of desired shapes \cite{crabbe1999second,kazadi04swarm,stewart06distributed,soleymani2014autonomous}.

The remainder of this paper is organized as follows.  Section II will discuss our hardware platform.  Section III will cover the algorithm and Section IV will focus on experimental results, both in simulation and on our physical robots.  Brief conclusions will follow in Section V.

\section{HARDWARE}

\begin{figure}[thpb]
    \centering
    \includegraphics[width=0.95\linewidth]{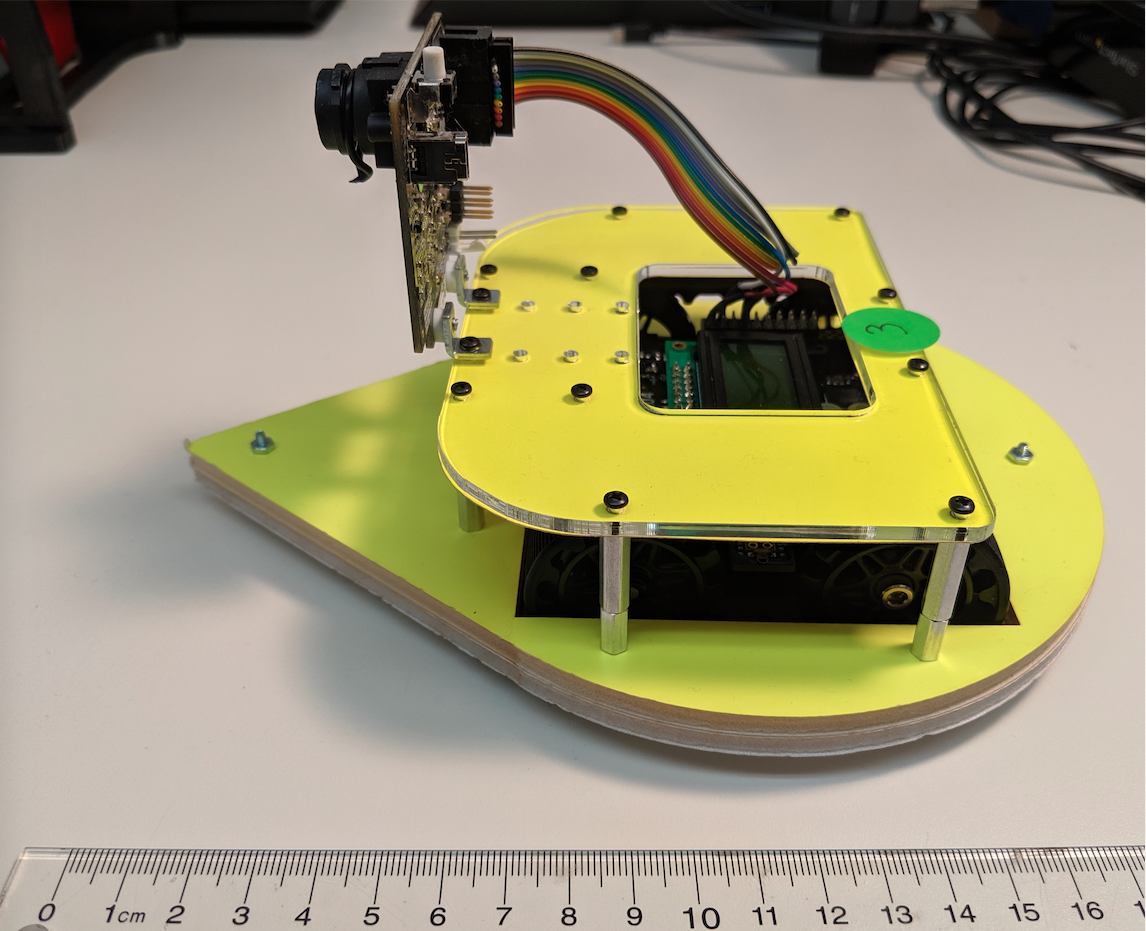}
    \vspace{0.25cm}

    \includegraphics[width=0.75\linewidth]{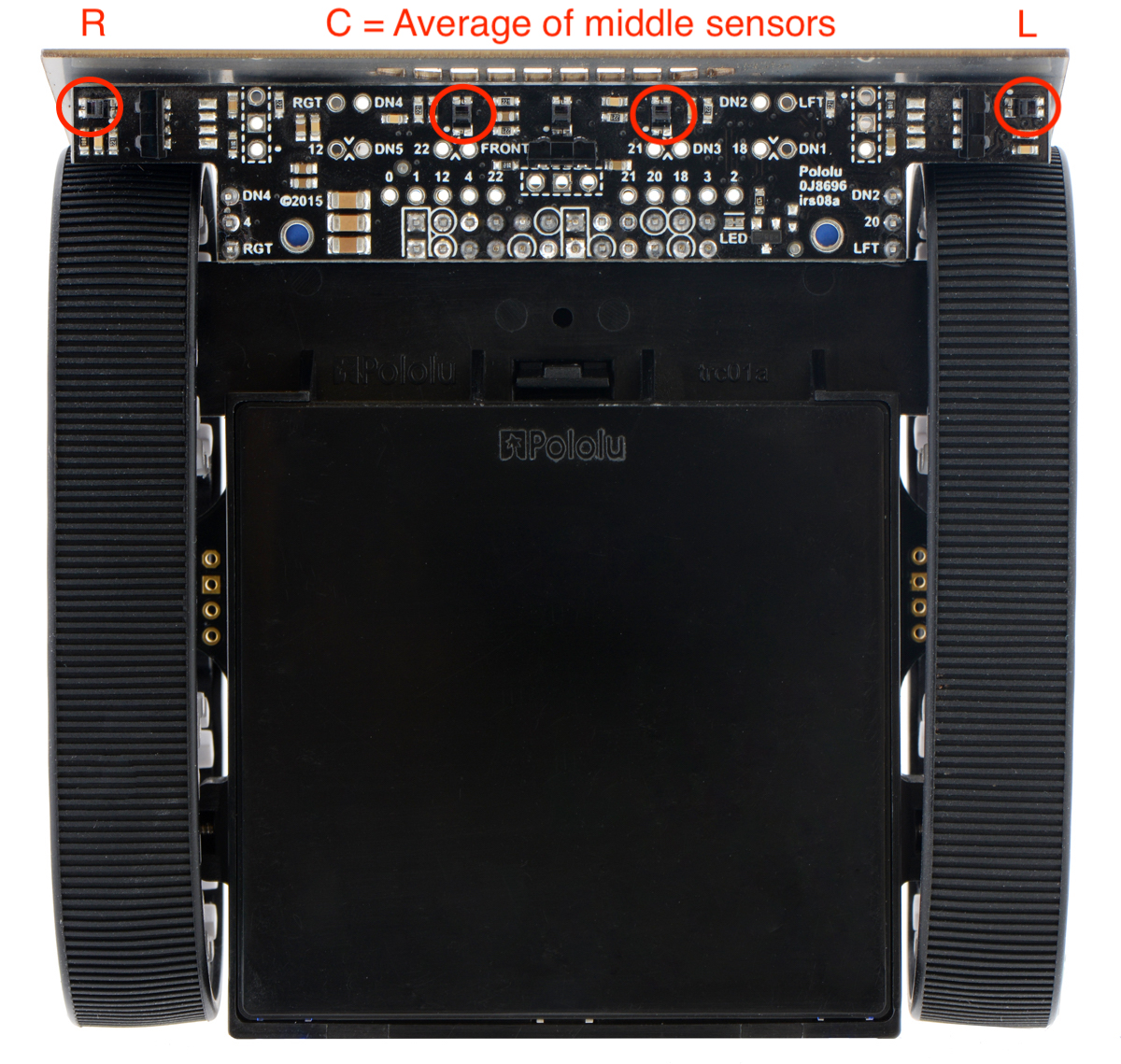}
    \caption{\emph{Top:} The robot.  \emph{Bottom:} Underside view of the Zumo32U4 with the 4 replaced reflectance sensors circled in red.  The sensors used to produce the values \LLL, \CCC and \RRR are also highlighted in red.}
    \label{fig:robot}
\end{figure}

Our robot, shown in Figure \ref{fig:robot}, is built upon the Zumo 32U4 robot, a $\approx$10 $\times$ 10 cm tracked differential-drive platform\footnote{\url{https://www.pololu.com/docs/0J63}}.  
The Zumo 32U4 is equipped with a linear array of 5 infrared reflectance sensors intended to detect black lines on white surfaces.  Each reflectance sensor consists of an infrared emitter/detector pair.  We modified 4 of these sensors (outlined in red in the bottom image of Figure \ref{fig:robot}) by replacing their infrared spectrum detectors with visible spectrum (600nm) phototransistors manufactured by Rohm Semiconductor (RPM-075PTT86).  As opposed to emitting light and sensing its reflection, our robots operate on a 75 inch diagonal LCD television manufactured by LG (75UK6190).  This television provides the light source for these phototransistors.  The image corresponding to the scalar field is projected on the television and then sensed.  Our algorithm requires sampling the scalar field at three points arranged in a triangle.  Since the physical sensors are co-linear we place the values obtained from the left and right sensors into a fixed-length queue to simulate 
sensors placed further back from the sensor array.  The middle two phototransistor values are averaged to produce the centre value of the triangle.  This approach is feasible because our robots generally move forwards, so older sensor values are similar to those that would appear to more posterior sensors.

\begin{figure}[thpb]
    \centering
    \includegraphics[width=0.95\linewidth]{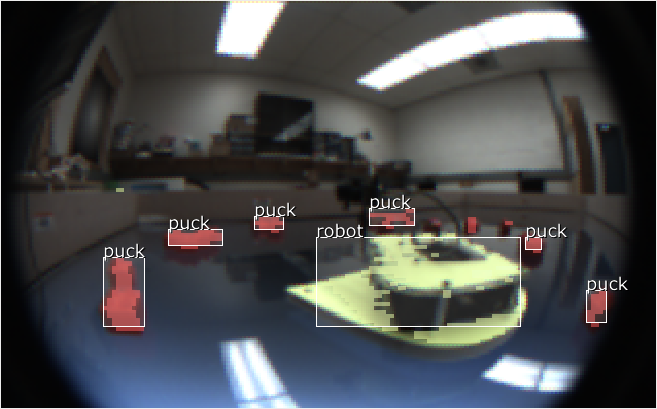}
    \caption{Image taken from the Pixy camera showing nearby pucks and another robot.  The annotated objects (`puck' and `robot') are those detected by the Pixy.}
    \label{fig:pixy}
\end{figure}

Connected to the Zumo is a Pixy\footnote{\url{https://pixycam.com/pixy-cmucam5/}} vision sensor which does on-board color segmentation and connected components labelling.  The Pixy produces a list of blocks, classified according to color as `puck' or `robot'.  
Figure \ref{fig:pixy} shows the Pixy's view of surrounding robots and pucks (red Lego pieces).  Note that a wide-angle lens has been installed on the Pixy.
% MAYBE NOT NEEDED
Our algorithm relies upon the detection of pucks or other robots which are known
to be on the left or right.  Rectangular blocks from the Pixy can easily be distinguished as being in the left or right half of the image.  Let $b_x$ indicate the $x-$coordinate of block $b$'s center, and let $b_w$ be the block's width.  If $W$ is the image width and $b_x - b_w/2 < W/2$ then the block is in the left half of the image.  If $b_x + b_w/2 > W/2$ then the block is in the right half.  A block may also straddle both sides of the image.  Note that we assume that the mid-point of the image at $W/2$ is aligned with the robot's centre which is the case for our robots.

The list of blocks detected by the Pixy is transmitted to the Zumo's ATmega32U4 microcontroller via I2C.  The current drawn by the Pixy is quite low (140 mA @ 5V) in comparison to general-purpose vision systems of a similar size and price point (e.g. a Raspberry Pi) allowing the whole system to be powered by the Zumo's on-board set of 4 standard AA batteries.% with endurance of ??? hours.  

Control code executes directly on the Zumo's microcontroller, the ATmega32U4.
Also note that around the robot's perimeter is a skirt made of foam board
attached to an acrylic plate mounted to the Zumo.  This skirt gives the robot a pointed wedge allowing it to extract objects that are adjacent to the sides and corners of the environment.

\section{ALGORITHM}

Our algorithm takes as input a set of sensor variables and produces the robot's forward speed $v$ and angular speed $\omega$.  The sensor variables are described below.

\begin{description}[labelsep=1em]
    \item[\LLL] (Scalar) Value of the scalar field as measured by the left sensor (simulated using a queue, as mentioned earlier).
    \item[\CCC] (Scalar) Value of the scalar field measured by the centre sensor.  \CCC is obtained by averaging the two middle sensors of the reflectance array (see Figure \ref{fig:robot}).
    \item[\RRR] (Scalar) Value of the scalar field as measured by the right
    sensor (simulated using a queue).
    \item[\leftPuck] (Boolean) Indicates the presence of pucks in the left half of the image (i.e. on the robot's left).
    \item[\leftRobot] (Boolean) Indicates the presence of other robots in the left half of the image.
    \item[\rightRobot] (Boolean) Indicates the presence of other robots in the right half of the image.
\end{description}

% PLACE FOR THIS?  MISSING DETAIL
%Note that the scalar field sensors nominally produce values in the range $[0, 1]$, but in practise our sensors' range is more limited than the output range of the LCD television.  

The algorithm (see Algorithm \ref{alg:oc2}) takes inspiration from Gauci \etal in reacting to pucks by moving forward while oscillating between veering left and right so that the puck's `outer' edge becomes the fixation point and the robot bumps into that edge, nudging the puck inwards towards the growing structure.  Whenever not reacting to a puck or another robot, it will guide the robot in a clockwise pattern around the structure.

A departure from the original algorithm is the way in which the desired shape is encoded in the scalar field.  Here the scalar field is set to zero to indicate a goal region where objects are to be collected.  In the original algorithm a particular threshold value of the field was sought, but the phototransistors used by our robots are sensitive to various noise sources and we found it more robust to sense the abrupt transition to zero as indicating the goal region.

The algorithm is applied on every time step and is reactive (i.e. stateless) with the exception of the recovery  action (lines 1-4) which is used to get the robot unstuck from the boundary or from other robots.  The algorithm's output is an ordered pair giving the forward and angular speeds of the robot.  Except when in the recovery state, these values are always chosen so that the robot moves forwards, veering to the left or right.

The original orbital construction algorithm mapped particular orderings of the
scalar field sensors into actions \cite{vardy2018orbital}, but this mapping was
not flexible.  In this version this mapping from orderings to actions is given
by the parameters \puckVariant and \alignVariantSTOP.  
Lines 5-16 determine the order of the three scalar field sensors, \LLL, \CCC and \RRR, which approximately capture the orientation of the robot with respect to the scalar field.  
Note that the conditions on lines 19 and 21 stipulate that the actions that follow on lines 20 and 22 (respectively) can only occur for certain values of the order variable---that is, certain orientations of the robot.  As an example, the condition on line 19 is active whenever the robot sees a puck on its left and then turns to collect it.  Yet, if the ordering is $\LLL \geq \RRR \land \RRR \geq \CCC$ this means the robot is turned away from the structure and oriented counter-clockwise.  If it tries to collect the puck it will strongly deviate from the nominal clockwise flow around the structure and perhaps get in the way of its peers.  The sixth bit of \puckVariant should be 0 to prevent this.  Similarly, the \alignVariant parameter determines for which orientations the robot will veer left or right if there is no puck on the left.  

The physical arrangement of sensors will dictate the best values to choose for
parameters \puckVariant and \alignVariantSTOP. 
We used the simulator described in the next section to exhaustively test all
64$^2$ values for both \puckVariant and \alignVariantSTOP.  We found that the values of \puckVariant = 13 and \alignVariant = 18 worked best for both single- or multiple-robot simulations.

\begin{figure}[thpb]
    \centering
    \includegraphics[width=0.95\linewidth]{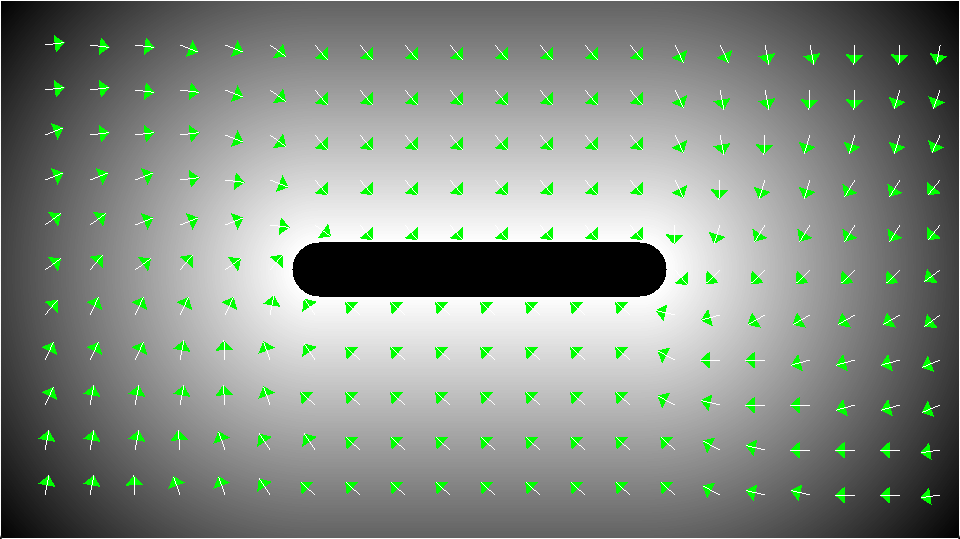}
    \vspace{0.25cm}

    \includegraphics[width=0.95\linewidth]{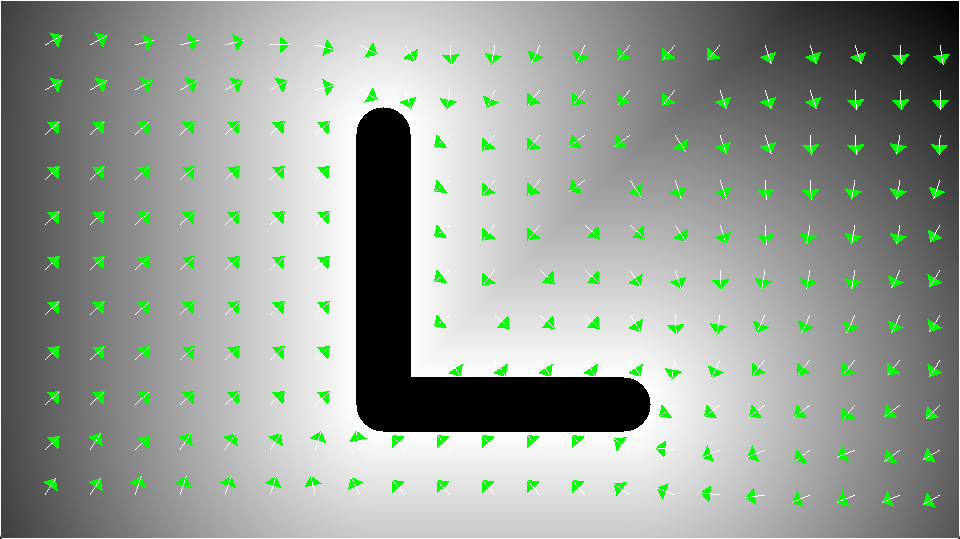}
    \caption{Vectors showing flow of robots in the absence of pucks or other robots for the line and `L' shapes.}
    \label{fig:vectors}
\end{figure}

Given the suitable settings for \puckVariant and \alignVariant the robots will be guided in a clockwise pattern around the structure.
Figure \ref{fig:vectors} shows vectors corresponding to how robots would move around two different shapes: a line segment and a letter `L'.

\begin{figure}
\removelatexerror
\begin{algorithm}[H]
    \SetKwInOut{Input}{input}
    \SetKwInOut{Output}{output}
    \SetKwInOut{Parameters}{params}
    \SetKwIF{If}{ElseIf}{Else}{if}{}{else if}{else}{end if}%

    \Input{\LLL, \CCC, \RRR, \leftPuck, \leftRobot, \rightRobot}
    \Output{(forward speed, angular speed)}
    \Parameters{\vMax, \omegaMax, \puckVariant, \alignVariant}
    \BlankLine

    \tcp{Handle getting stuck}
    \If {scalar field sensors not changing}{
        set recovery timer\;
    }
    \If {recovery timer not elapsed}{
        \Return{ randomized reverse $(v, \omega)$}\;
    }
    \BlankLine

    \tcp{Set order as a binary number}
    \uIf {$\RRR \geq \CCC \land \CCC \geq \LLL$}{
       order = 0b000001\;
    }
    \uElseIf {$\CCC \geq \RRR \land \RRR \geq \LLL$}{
        order = 0b000010\;
    }
    \uElseIf {$\RRR \geq \LLL \land \LLL \geq \CCC$}{
        order = 0b000100\;
    }
    \uElseIf {$\LLL \geq \RRR \land \RRR \geq \CCC$}{
        order = 0b001000\;
    }
    \uElseIf {$\CCC \geq \LLL \land \LLL \geq \RRR$}{
        order = 0b010000\;
    }
    \ElseIf {$\LLL \geq \CCC \land \CCC \geq \RRR$}{
        order = 0b100000\;
    }
    \BlankLine

    \tcp{Choose action}
    \uIf {\RRR indicates black and $\CCC \geq \LLL$}{
        \tcp{Veer left away from goal zone}
        \Return{$(\vMax, \omegaMax)$}\;
    }
    \uElseIf {$(\puckVariant \mathbin{\&} order \ne 0) \land \leftPuck 
           \land \lnot \leftRobot$}{
        \tcp{Veer left to gather puck}
        \Return{$(\vMax, \omegaMax)$}\;
    }
    \uElseIf {$(\alignVariant \mathbin{\&} order \ne 0) \land \lnot
               \leftRobot$}{
        \tcp{Veer left to align with scalar field}
        \Return{$(\vMax, \omegaMax)$}\;
    }
    \ElseIf {$\lnot \rightRobot$}{
        \tcp{Veer right}
        \Return{$(\vMax, -\omegaMax)$}\;
    }
    \BlankLine

    \tcp{Go slowly forwards}
    \Return{$(0.25 \vMax, 0)$}\;

    \caption{Orbital Construction 2.0}
    \label{alg:oc2}
\end{algorithm}
\end{figure}

\section{RESULTS}

\subsection{Simulation}

\begin{figure}[thpb]
    \centering
    \includegraphics[width=1.0\linewidth]{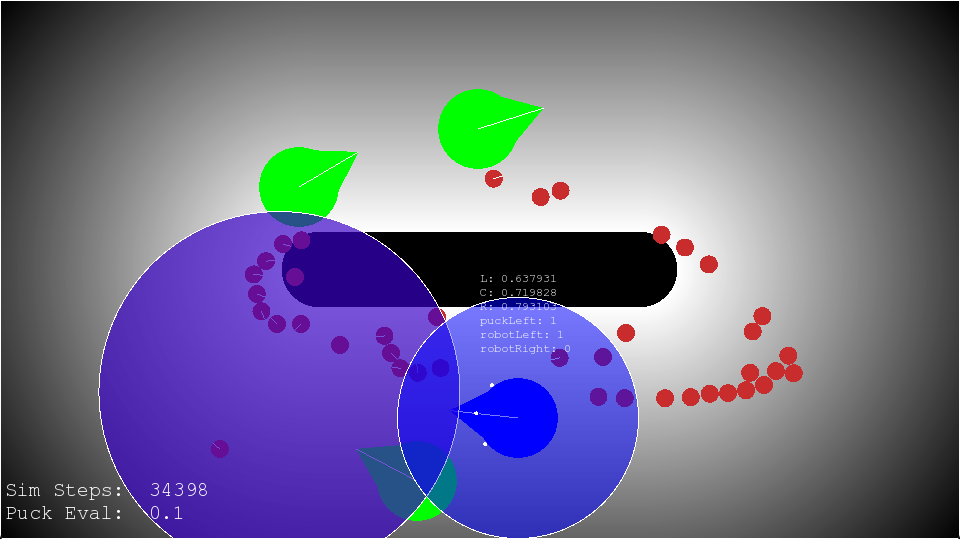}
    \caption{Screenshot from our simulator with 4 robots constructing a line segment.}
    \label{fig:cwaggle}
\end{figure}

We have developed a custom simulation environment\footnote{\url{https://github.com/BOTSlab/cwaggle/tree/orbital_av}} to predict how the algorithm will perform in different situations.  This simulator is a fork of the cwaggle simulator\footnote{\url{https://github.com/davechurchill/cwaggle}} developed by David Churchill which achieves a very high update rate by a design that optimizes cache coherency and simulates almost all bodies and sensors as circles.  Figure \ref{fig:cwaggle} presents a screenshot.  The blue robot is highlighted and the circular regions representing its sensors are shown.  The larger circular region represents the field-of-view of its camera.  This field-of-view is restricted by the smaller circle for sensing other robots. 

At the moment shown in Figure \ref{fig:cwaggle} the blue robot senses the puck on the bottom left.  The robot would normally turn left to collect this puck, but it also senses the other robot on its left.  So it will go slowly forwards as dictated by line 25 of the algorithm.

\begin{figure}[thpb]
    \centering
    \includegraphics[width=1.0\linewidth]{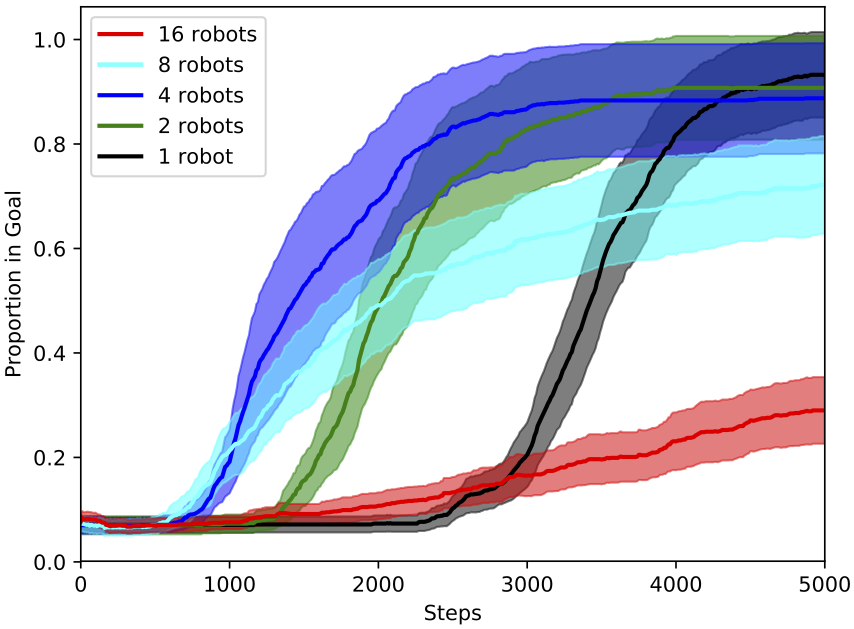}
    \caption{Performance in simulator on the line shape when the number of robots is varied.  Each trace is an average of 30 trials.  Shaded regions with matched colors represent 95\% confidence intervals.}
    \label{fig:sim}
\end{figure}

This simulator allows us to make predictions about performance in advance of hardware experiments.  It is naturally important to understand how the algorithm performs when the number of robots is varied.  Figure \ref{fig:sim} shows this variation over 30 trials for each number of robots applied to the line segment shape with 40 pucks present.  We measure performance by the proportion of pucks touching the goal region.  With 1 robot, a high value is eventually reached, but with 2 or 4 robots a similar proportion is reached much more quickly.   However, with 8 or 16 robots the performance suffers as the degree of spatial interference increases.

\begin{figure}[thpb]
    \centering
    \includegraphics[width=0.95\linewidth]{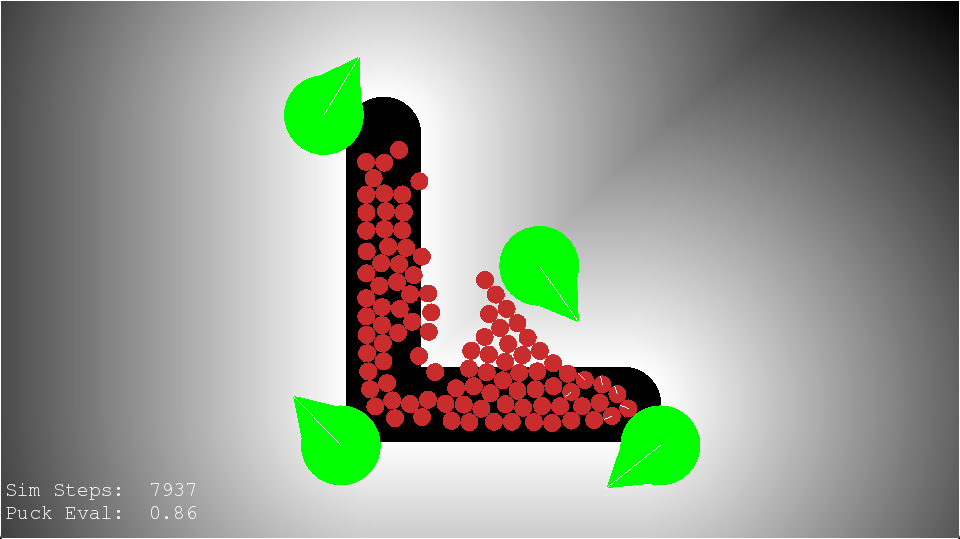}
    \vspace{0.25cm}

    \includegraphics[width=0.95\linewidth]{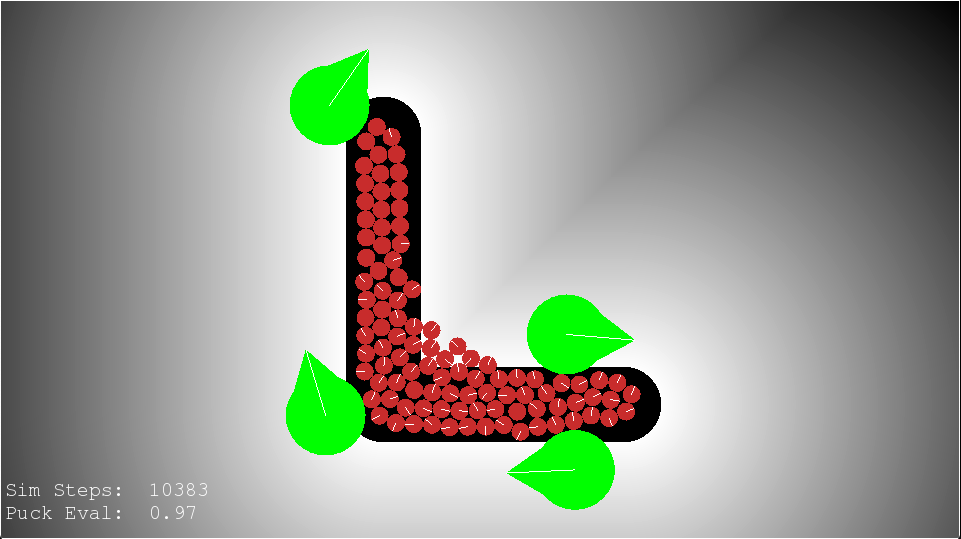}
    \vspace{0.25cm}

    \includegraphics[width=0.95\linewidth]{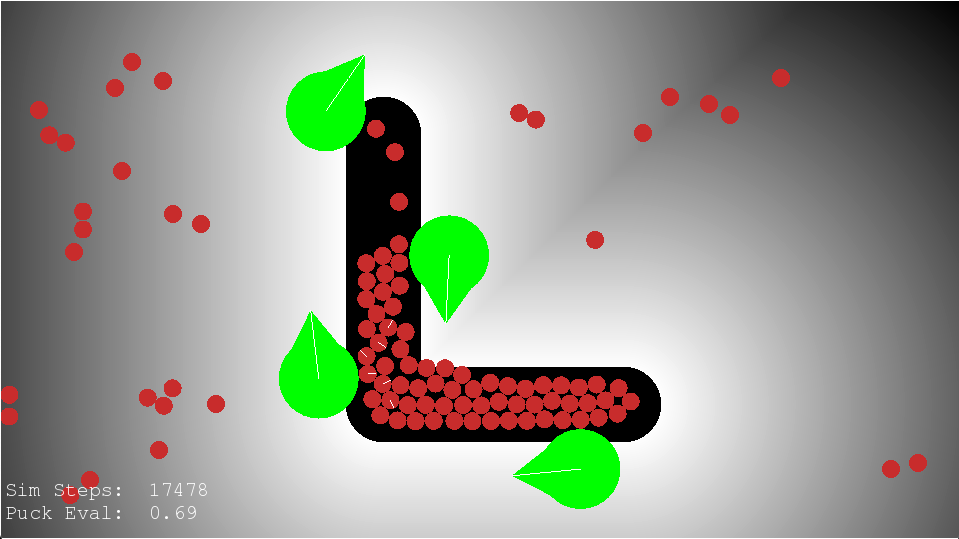}
    \caption{Simulation result on the `L' shape with the puck-sensing radius set to 420 (top), 100 (middle) and 80 (bottom).}
    \label{fig:sim_L}
\end{figure}

To probe the variety of shapes that can be produced we tested an `L' shape.  The concavity of this shape presents somewhat of a challenge.  The top image in Figure \ref{fig:sim_L} shows the converged shape when the radius of puck sensing is set to the default value of 420 (the simulator uses units of pixels).  Clearly the interior of the shape is not well-formed.  This is because the algorithm reacts to any puck sensed on its left, so any pucks on the bottom-right of the `L' will always draw the robot away from the interior.  We can address this by reducing the radius within which pucks are detected to 100, which is shown in the middle image of Figure \ref{fig:sim_L}.  Here the shape is much more accurate.  However, if the puck-sensing radius is reduced further to 80 then the robots lose the ability to gather all pucks into the structure as shown in the bottom image.

\begin{figure}[thpb]
    \centering
    \includegraphics[width=1.0\linewidth]{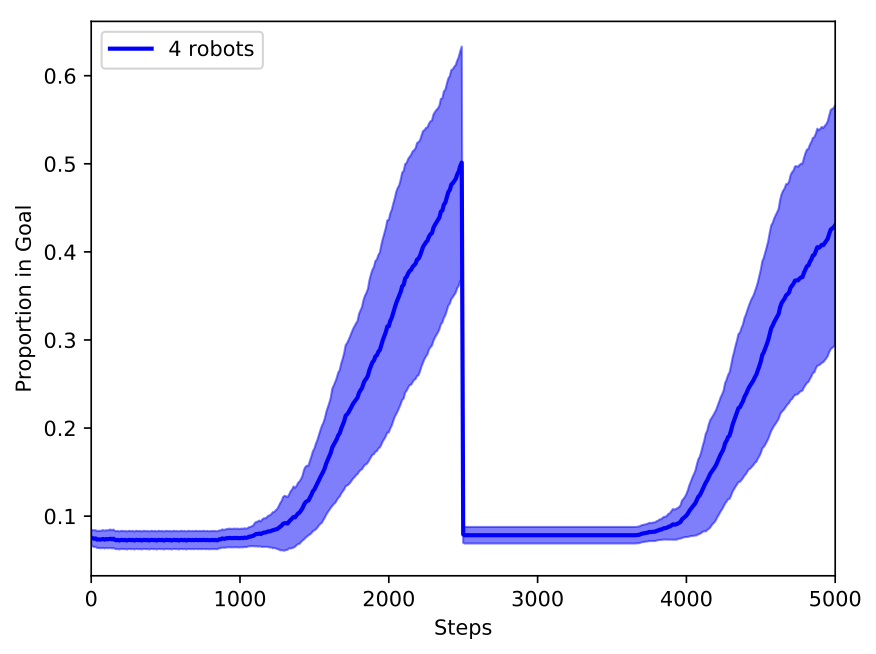}
    \caption{Performance in simulator on the line shape when all pucks are randomly repositioned at time step 2500.  Conditions as per Figure \ref{fig:sim}.}
    \label{fig:reposition}
\end{figure}

Another interesting aspect is the ability of our method to withstand
perturbations.  In trials involving 4 robots, all pucks were randomly repositioned mid-trial (time step 2500).  Figure \ref{fig:reposition} shows that the system manages to recover from this perturbation and regain a similar threshold of performance.

\subsection{Physical Robots}

We completed trials on our physical robots using a scalar field image specifying a line segment.  The robots and 40 square pieces of red Lego were randomly positioned within the arena at the start of each trial.  These items were placed at least 5 cm away from the boundary of the environment.
Each trial was executed for 4 minutes and judged based on the number of pucks touching the goal region at the end of the trial.  For each number of robots the trial was repeated 3 times.  No claims of statistical significance can be drawn but these results do provide a useful qualitative assessment.  A summary of these trials is found in table \ref{tab:results}. Note that the second column in table \ref{tab:results} refers to the number of pucks within or touching the goal region at the end of the trial.

\begin{table}[thpb]
    \tabulinesep=1.05mm
    \begin{tabu}{|l|l|X|}
    \hline
    \textbf{Robots} & \textbf{\#} & \textbf{Notes} \\\hline
    1 & 39 & - \\
    1 & 36 & - \\
    1 & 31 & - \\\hline
    2 & 40 & - \\
    2 & 36 & Robots become stuck and unstuck \\
    2 & 40 & - \\\hline
    3 & 36 & 2 robots stuck together throughout majority of run, trapping 4 pucks \\
    3 & 40 & - \\
    3 & 13 & 3 robots stuck together during later half of run, trapping 1 puck \\\hline
    4 & 6 & various combinations of robots stuck together throughout run \\
    4 & 40 & various combinations stuck;  all but one recover \\
    4 & 39 & various combinations stuck; all but one recover \\\hline
    \end{tabu}
    \caption{Summary of hardware trials.}
    \label{tab:results}
\end{table}

\begin{figure}[thpb]
    \centering
    \includegraphics[width=0.95\linewidth]{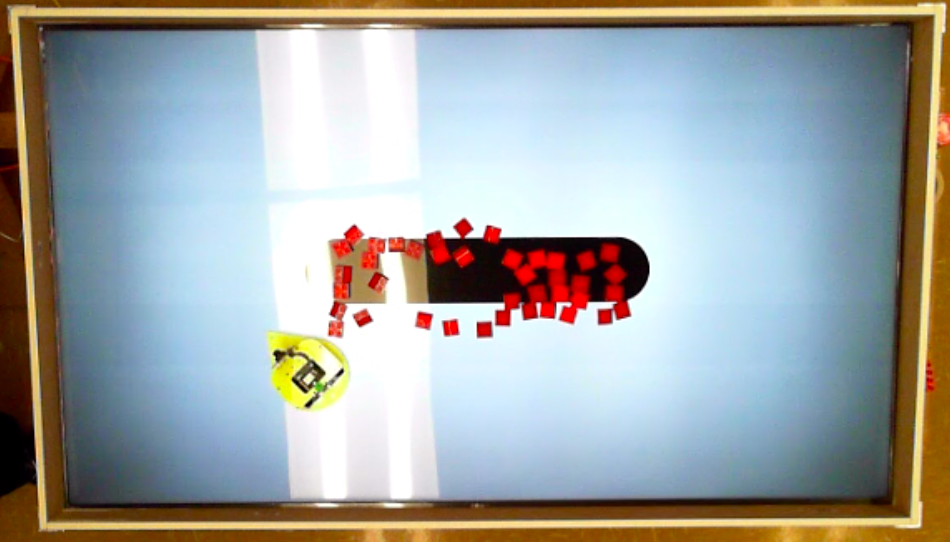}
    \vspace{0.25cm}

    \includegraphics[width=0.95\linewidth]{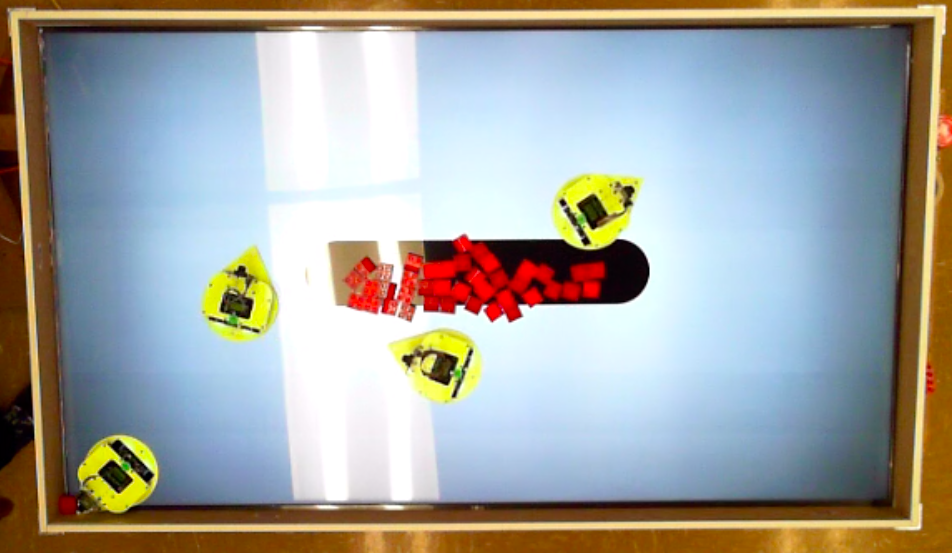}
    \caption{Final result from hardware trials for 1 robot (top) and 4 robots (bottom).  These are from the third trial listed for each number of robots as given in Table \ref{tab:results}.  The large white vertical bands are reflections of the overhead lights.}
    \label{fig:trials}
\end{figure}

Figure \ref{fig:trials} shows the final result for the third trial with 1 robot and with 4 robots.  Ideally all 40 pucks would be touching the goal region but the trial shown in Figure \ref{fig:trials} (top) shows 9 pucks lying wholly outside this region, although they remain quite close.  This pattern is consistent within the hardware trials but was not seen in the simulation results.  We believe it indicates a blemish or irregularity in the projected scalar field image.   Figure \ref{fig:trials} (bottom) shows a more successful final result in that 39 pucks lie within the goal region, with one captured by the robot stuck in the corner.

\section{CONCLUSIONS}

We have presented a hardware platform and a new algorithm for the planar construction problem.  Our experimental results are promising but it is clear that interference between robots is impeding performance.  We attempted several different approaches to avoid collisions with other robots, but this remains a challenge for reactive robot controllers such as ours.  However, we have demonstrated a feasible platform for future research in this direction and have shown that readily available components such as the Zumo, Pixy, phototransistors and a television can be combined into a working swarm robotic system to investigate solutions to the planar construction problem.

\addtolength{\textheight}{-12cm}   % This command serves to balance the column lengths

%%%%%%%%%%%%%%%%%%%%%%%%%%%%%%%%%%%%%%%%%%%%%%%%%%%%%%%%%%%%%%%%%%%%%%%%%%%%%%%%
%\section*{ACKNOWLEDGMENT}
%%%%%%%%%%%%%%%%%%%%%%%%%%%%%%%%%%%%%%%%%%%%%%%%%%%%%%%%%%%%%%%%%%%%%%%%%%%%%%%%

%\bibliographystyle{IEEEconf}
%\bibliography{IEEEconf,/Users/av/av_work/doc/refs}
\bibliographystyle{IEEEtran}
\bibliography{refs}

\end{document}